\documentclass{article}

\usepackage{arxiv}

\usepackage[utf8]{inputenc} 
\usepackage[T1]{fontenc}    
\usepackage{hyperref}       
\usepackage{url}            
\usepackage{booktabs}       
\usepackage{amsfonts}       
\usepackage{nicefrac}       
\usepackage{microtype}      
\usepackage{lipsum}		
\usepackage{graphicx}
\usepackage{doi}
\usepackage{subcaption}
\usepackage{amsmath}
\usepackage{multirow}

\title{Transferring Graph Neural Networks for Soft Sensor Modeling using Process Topologies}

\author{ \href{https://orcid.org/0009-0008-7832-6840}{\includegraphics[scale=0.06]{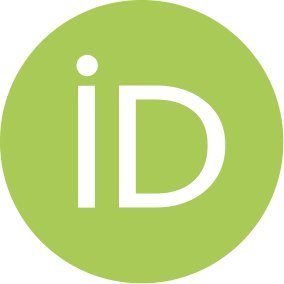}\hspace{1mm}Maximilian F. Theisen} \\
    Process Intelligence Research \\
	Department of Chemical Engineering\\
	Delft University of Technology\\
	Delft, 2629 HZ, Netherlands \\
    \And
	\hspace{1mm}Gabrie M. H. Meesters \\
	Department of Chemical Engineering\\
	Delft University of Technology\\
	Delft, 2629 HZ, Netherlands \\
	\And
	\href{https://orcid.org/0000-0001-8885-6847}{\includegraphics[scale=0.06]{figures/orcid.png}\hspace{1mm}Artur M. Schweidtmann} \\
    Process Intelligence Research \\
	Department of Chemical Engineering\\
	Delft University of Technology\\
	Delft, 2629 HZ, Netherlands \\
	\texttt{a.schweidtmann@tudelf.nl} \\
}

\renewcommand{\shorttitle}{\textit{Transfer learning for soft sensors between process topologies}}

\hypersetup{
pdftitle={Transfer learning for soft sensors},
pdfsubject={soft sensors},
pdfauthor={Maximilian F. Theisen},
pdfkeywords={Digital twins, Machine learning, Process operations, Graph neural networks, Transfer learning},
}

\begin{document}
\maketitle

\begin{abstract}
Data-driven soft sensors help in process operations by providing real-time estimates of otherwise hard-to-measure process quantities, e.g., viscosities or product concentrations.  
Currently, soft sensors need to be developed  individually per plant.  
Using transfer learning, machine learning-based soft sensors could be reused and fine-tuned across plants and applications.  
However, transferring data-driven soft sensor models is in practice often not possible, because the fixed input structure of standard soft sensor models prohibits transfer if, e.g., the sensor information is not identical in all plants.  
We propose a topology-aware graph neural network approach for transfer learning of soft sensor models across multiple plants.  
In our method, plants are modeled as graphs: Unit operations are nodes, streams are edges, and sensors are embedded as attributes.  
Our approach brings two advantages for transfer learning:  
First, we not only include sensor data but also crucial information on the plant topology.  
Second, the graph neural network algorithm is flexible with respect to its sensor inputs.  
This allows us to model data from different plants with different sensor networks.  
We test the transfer learning capabilities of our modeling approach on ammonia synthesis loops with different process topologies~\cite{Araujo2008Controlstructuredesign}.  
We build a soft sensor predicting the ammonia concentration in the product.  
After training on data from one process, we successfully transfer our soft sensor model to a previously unseen process with a different topology.  
Our approach promises to extend the data-driven soft sensors to cases to leverage data from multiple plants.  

\end{abstract}


\section{Introduction}
\label{intro:esapce_tl}
Data-driven soft sensors promise to increase operating efficiency in chemical plants by providing real-time estimates of hard-to-measure quantities.  
Previous works have shown the merit of soft sensors in many applications, e.g., estimating viscosity or product concentrations~\cite{Liu2019Domainadaptationtransfer, Bispo2017DevelopmentANNbased}.  
However, data scarcity remains a major hindrance in developing machine learning-based solutions such as soft sensors in industry~\cite{Bortz2023AIProcessIndustries}.  

Transfer learning offers an opportunity to decrease data requirements for soft sensor modeling but is often not possible for processes with different topologies.  
In transfer learning, a machine learning model pretrained on one domain is transferred to another domain.  
Transfer learning for soft sensors has shown several advantages, such as reduced data requirements, zero-shot capabilities, and improved performance~\cite{Curreri2021SoftSensorTransferability}.  
In previous literature, transfer learning has been explored in multiple situations for reducing data requirements in soft sensor development.  
For domain adaptation, the model transfers knowledge between different operating points~\cite{Liu2019Domainadaptationtransfer}.  
Similarly, transfer learning has been leveraged to transfer between batch runs~\cite{Kay2024Integratingtransferlearning}.  
It has even been applied to transfer between identical process lines~\cite{Curreri2021RNNLSTMBased, Farahani2021DomainAdversarialNeural}.  
While different types of models have been used in these works (multi-layer perceptrons (MLPs), extreme learning machines, recurrent neural networks), it is not possible to transfer these soft sensor models between plants with different topologies, e.g., with respect with to sensor locations or equipment types, due to the fixed input structure of the models.

We propose a new soft sensor architecture that uses spatio-temporal graph neural networks (GNNs) on the process topology and sensor data.  
This enables transfer learning between different processes due to its flexible input structure and topology-awareness.  
We model chemical processes as graphs, with unit operations as nodes, streams as edges, and sensor data embedded as attributes.  
Our approach allows to incorporate information about the process topology along with the sensor measurements.  
We argue that this approach has two advantages for transfer learning:  
First, this data representation provides context of the process to the model.  
This enables the model to learn about the process and thus transfer information to related processes.  
Second, our graph-based model does not have a fixed input size, meaning that it can be transferred between processes with different sensor networks.  
We demonstrate our modeling approach on two ammonia synthesis loops.  
The processes are similar, but topologically different, making them challenging for standard transfer learning approaches.  

\section{Methods}
\label{methods:escape_TL}

\subsection{Spatio-temporal modeling of soft sensors with graphs}
\begin{figure}
    \centering
    \includegraphics[width=1\linewidth]{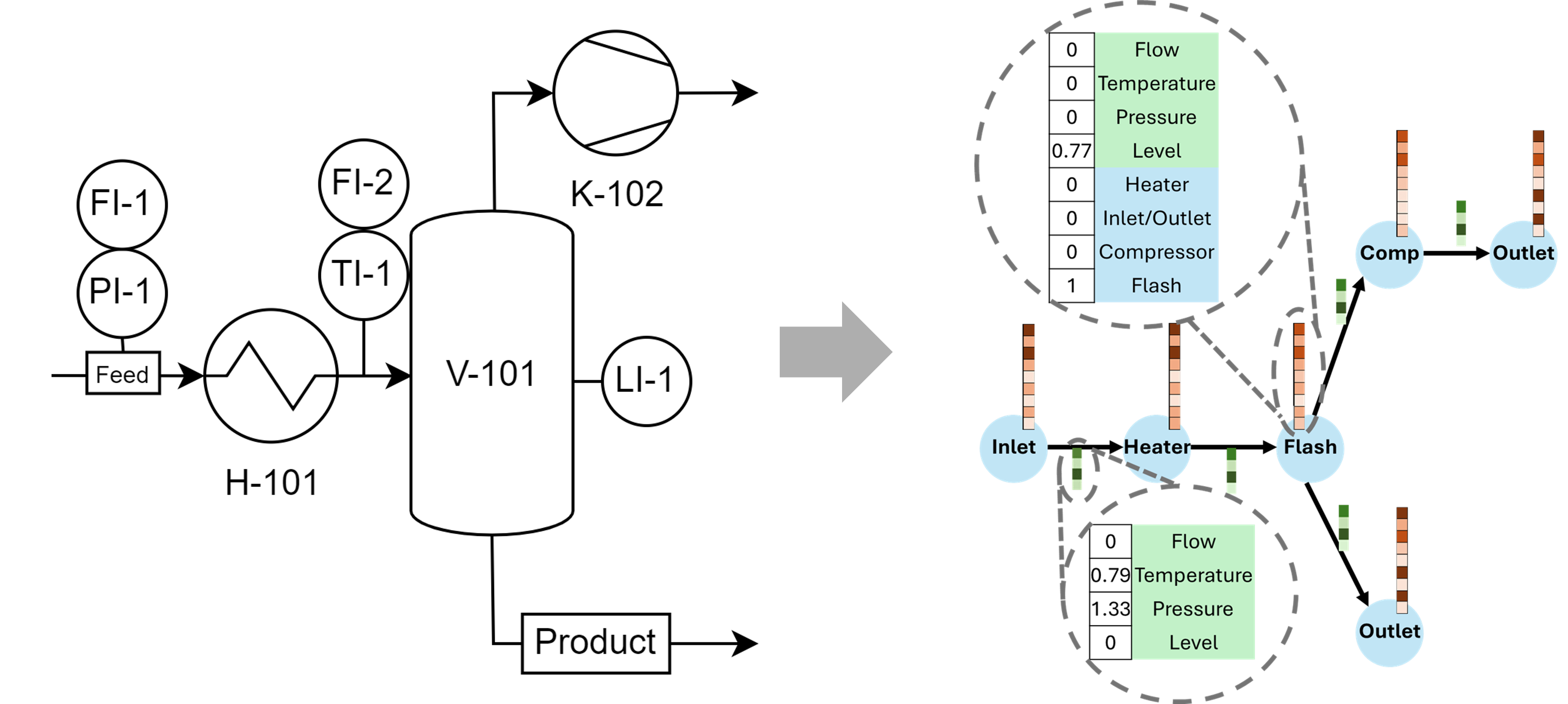}
    \caption{Visualization of flowsheet graph representation of an illustrative flowsheet.}
    \label{fig:graph-rep}
\end{figure}

To enable transfer learning between processes, we utilize a graph-based representation of the process topology and the soft sensor data to enable topology-awareness together with a model that accounts for both spatial and temporal dependencies in the data.  
We represent the underlying process as a directed graph, representing the topology of the process, see Figure~\ref{fig:graph-rep}.  
We model unit operations as nodes.  
To model different types of units, we one-hot encode the unit type into the node attribute vector.  
We further represent streams as directed edges.  
The direction of the edge follows the direction of the material.  
Finally, we encode sensor measurements according to their location in the process into the attribute vectors of nodes and edges.  

\begin{figure}
    \centering
    \includegraphics[width=1\linewidth]{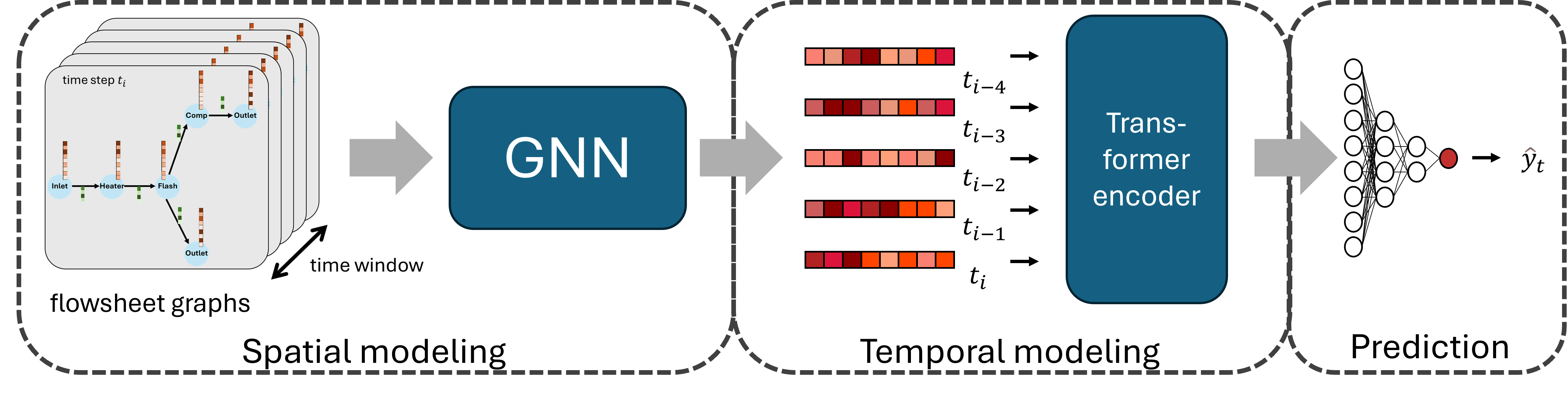}
    \caption{Illustration of the model used for spatio-temporal modeling, here with a look back window of 5 time steps.}
    \label{fig:spatio-temp-GNN}
\end{figure}

A complete overview of the modeling framework can be found in Figure~\ref{fig:spatio-temp-GNN}.  
We model the spatial relationships between the graph-encoded data and the soft sensor target using GNNs for each time step.  
We deploy a message-passing GNN~\cite{Gilmer2017Neuralmessagepassinga}, allowing us to utilize both node and edge information.  
The output is an individual flowsheet embedding for each time step, see Figure~\ref{fig:spatio-temp-GNN}.  
To model the temporal relationship and thus dynamics of the process, we use a transformer.  
The transformer encoder inputs the graph-encoded flowsheet embeddings.  
In its final layer, the encoder averages over the latent space along the number of time steps, resulting in a single output vector.  
A three-layer MLP then predicts the sensor target $\hat{y}_t$ for the current time step $y$.

\subsection{Transfer learning}
The transfer learning process in this work is conducted in two stages.  
First, the model undergoes initial training on the source domain.  
Subsequently, a partial retraining is performed on the target domain.  
We also test the case of no retraining on the target domain, which is referred to as zero-shot transfer learning.  

To further stabilize the training, we scale our input data by applying a log-scale.  
Since each process has a different topology and set of embedded sensor measurements due to different sensor information, different streams, etc., we cannot uniformly apply per-feature normalization across them.  
We, however, found that with sufficiently small learning rates, the training is still stable.  
We do normalize our soft sensor target to zero mean and unit standard deviation.  

We evaluate the model prediction $\hat{y}$ against the ground truth $y$ on our test set $N_\text{test}$ for all time steps $t$ using the root mean squared error (RMSE) on the normalized data:  

\begin{equation}
\text{RMSE} = \sqrt{\frac{1}{N_\text{test}} \sum_{t=1}^{N_\text{test}} \left( y_t - \hat{y}_t \right)^2}
\end{equation}

\section{Case study: Ammonia synthesis loops}
\label{cs:ammonia_syntesis}
We consider two ammonia synthesis loops as illustrative case studies with the same equipment sizing and feed flows but different topologies.  
In both processes, ammonia is produced from N\textsubscript{2} and H\textsubscript{2} under high pressure.  
The soft sensor target is the concentration of ammonia in the respective product streams.  
Both processes consist of four types of processing steps: (1) Compression, (2) Separation via flashing, (3) Heating/Cooling, and (4) Reaction with three reactor beds.  
The combination of these major units, however, differs between the two processes, reflecting different topologies found in industry~\cite{Moulijn2013Chemicalprocesstechnology}.  

\begin{figure}
    \centering
    \includegraphics[width=1\linewidth]{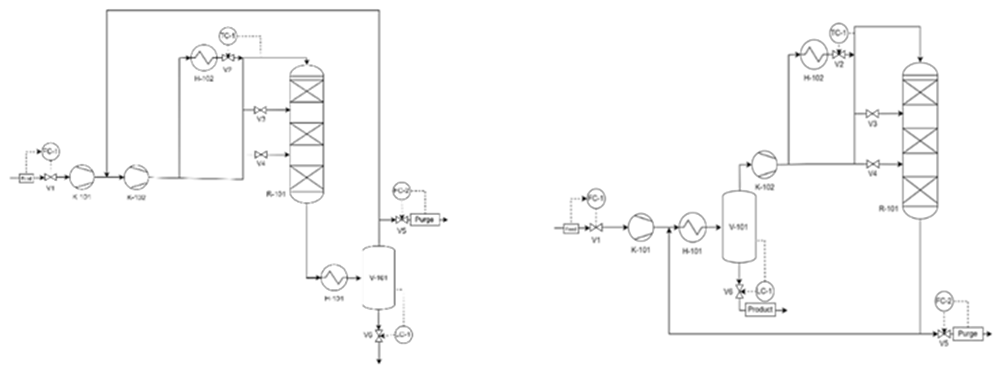}
    \caption{Two ammonia synthesis loops are used in this work. In Process A (left) the feed is first sent to the reactor and then flashed. In Process B, the feed is first flashed and then sent to the reactor.}
    \label{fig:two-ammonia-loops}
\end{figure}

In Process A (see Figure~\ref{fig:two-ammonia-loops}, left), the feed is first compressed in compressor K-101 and mixed with the recycle.  
The mixture is then further compressed in compressor K-102 and heated before entering the reactor R-101.  
The effluent leaving the reactor is flashed in the flash vessel V-101.  
The liquid phase leaving V-101 is the product stream, while the vapor phase of V-101 is purged and mixed with the inlet gas.  

In Process B (see Figure~\ref{fig:two-ammonia-loops}, right), the feed is first compressed in compressor K-101 and combined with the reactor outlet.  
The mixture is then cooled and flashed in V-101, removing the ammonia as a liquid.  
The liquid ammonia then leaves the process as its product.  
The vapor phase leaving the flash V-101 is further compressed in compressor K-102 and heated, after which it is sent into the reactor R-101.  
The reactor outlet is purged before being mixed with the inlet stream.  

Both processes are controlled using PID control.  
The control scheme follows the “Mode I” as detailed by~\cite{Araujo2008Controlstructuredesign}.  
Both the feed and the purge stream are controlled with a flow controller (FC-1 and FC-2, respectively).  
The level in the flash vessels is controlled with a level controller (LC-1).  
The reactor inlet temperature is controlled with the temperature controller (TC-1).  

We simulate dynamic operating datasets from the processes using the dynamic simulation software Aspen Plus Dynamics V12.  
As a design basis, we use the simulation files by Araújo and Skogestad~\cite{Araujo2008Controlstructuredesign}.  
To generate a dataset of dynamic operation, we randomly vary the setpoints of controllers in a range of 1–20\% deviation from the original, running the simulation until steady state is reached on the new setpoint.  
Then, we revert the setpoint back to the original.  
This way, we simulate 80 hours of operation.  
We sample sensor data every 36 seconds.  
This results in approximately 8000 data points for both datasets.  
We split the dataset 80/20/20 between training, validation, and testing in a chronological split.  

\section{Results}
\label{results-Escape-TL}
We carry out the transfer learning with our topology-aware GNN in a two-step process:  
(1) We first train our model on the dataset of Process A until we reach convergence on our validation set.  
(2) We then test the transfer capabilities of the GNN model on Process B.  

For this, we fine-tune the pretrained model on fractions of Process B, specifically on up to 51 points.  
This low number of training points reflects the often limited number of soft sensor target measurements available in industrial settings.  
For comparison, we also train a model on the data from Process B from scratch.  
We additionally test the zero-shot capability of the model pretrained on Process A.  
We train each model nine times using different seeds to mitigate the effects of training noise, and plot both the mean and standard deviation of the nine models combined.  
For training, we used a NVIDIA GeFORCE RTX 3090.  
The training times varied between 30 minutes for the full training dataset and 5 minutes for a fraction of the dataset.  

\begin{figure}
    \centering
    \includegraphics[width=0.5\linewidth]{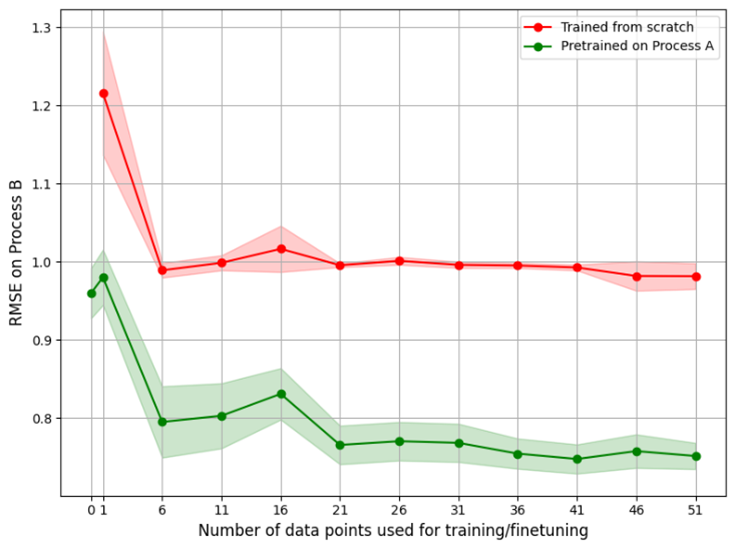}
    \caption{Pretrained vs training from scratch for varying number of datapoints. The shaded area shows the standard deviation due to different weight initializations}
    \label{fig:results-scratch-vs-finetune}
\end{figure}

The results are shown in Figure~\ref{fig:results-scratch-vs-finetune}.  
We plot the fraction of the data from Process B used for training vs. RMSE on the test set of Process B.  
We also illustrate the zero-shot capabilities of the pretrained model by adding a data point for 0 datapoints used for training.  

By analyzing the results shown in Figure~\ref{fig:results-scratch-vs-finetune}, we observe three notable trends: 
(1) The zero-shot performance of the pretrained model is significant, achieving an RMSE of 0.9753 on Process B without any training data from Process B.  
In contrast, the model trained from scratch requires 46 datapoints to reach a similar performance level.  
We attribute this to the pretrained model's ability to transfer learned dynamics from Process A to the previously unseen Process B.  
(2) Both models improve their RMSE scores on Process B when trained on more data.  
The pretrained model achieves an RMSE of 0.7633 when trained on 51 datapoints, while the model trained from scratch also shows improvement, reaching an RMSE of 0.9776.  
However, the pretrained model consistently outperforms the model trained from scratch across different training data sizes.  
(3) The performance of the pretrained model deteriorates when fine-tuned on a single datapoint from Process B, performing worse than its zero-shot performance.  
We hypothesize that this might be due to the lack of sufficient information in a single datapoint, which could lead to overfitting or misrepresentation of the underlying process dynamics.  

\begin{table}[ht!]
\centering
\caption{Average percentage RMSE reduction when using the pretrained soft sensor model compared to the model trained from scratch on Process B.}
\begin{tabular}{|c|c|c|c|c|c|c|}
\hline
Number of datapoints used for training on Process B & 1   & 11  & 21  & 31  & 41  & 51  \\ \hline
Average percentage RMSE reduction                  & 19.38\% & 19.36\% & 23.08\% & 22.80\% & 24.15\% & 23.87\% \\ \hline
\end{tabular}
\label{tab:results-escape-tl-rmse-reduction}
\end{table}

The relative reduction in RMSE when using the soft sensor model pretrained on Process A compared to the soft sensor model trained from scratch is further calculated in Table~\ref{tab:results-escape-tl-rmse-reduction}, showing up to 24.15\% in reduction. 
It is notable that the relative reduction in RMSE increases with more training data from Process B. We thus hypothesize that the model pretrained on Process A may benefit more from additional training data of process B than the model trained from scratch in this limited data regime.

\begin{figure}
    \centering
    \includegraphics[width=1\linewidth]{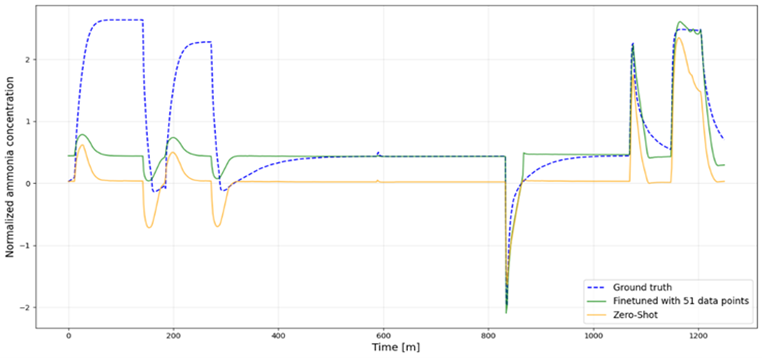}
    \caption{Visualization of test set (blue) vs fine-tuned model (orange, green). We visualize both the zero-shot performance (orange) and the model fine-tuned on 51 datapoints.}
    \label{fig:results-scratch-vs-finetune-over-time}
\end{figure}

We further visualize the effect of fine-tuning in Figure~\ref{fig:results-scratch-vs-finetune-over-time}.  
We visualize the ground truth alongside the zero-shot predictions from the pretrained model and the predictions from the fine-tuned model over time on the Process B test set.  
The zero-shot model underpredicts the ammonia concentration consistently.  
This is even the case during steady state, e.g., between minute 500 and 800.  
This can be explained by the fact that the ammonia concentration in Process A and Process B are not identical, and thus the model still predicts an ammonia concentration as it would occur in Process A.  
With fine-tuning, this offset is reduced, as shown by the prediction of the fine-tuned model between minutes 500 and 800.  
Nevertheless, some of the dynamics, e.g., between minute 0 and 350, are still not well captured by the fine-tuned model.  

\section{Conclusion}
We propose a spatio-temporal soft sensor modeling framework using GNNs to enable transfer learning across topologically different processes. 
We utilize two similar but topologically different ammonia synthesis loops as our case study. We show that the topology-aware GNN is transferred from one process to the other. 
We further demonstrate how transfer learning reduces data requirement through good performance without any data on the target domain and even better performance on very little additional data. We show that using the pretrained model, we can reduce the RMSE by up to 24.15\% when trained on 46 datapoints compared to the model trained from scratch.

This work presents a flexible framework that can be extended to various chemical process systems.  
Future work could focus on adapting this transfer learning framework to multi-process environments, generalizing beyond one-process pretraining.  
Further, the approach is yet to be verified on industrial data, which often poses additional challenges.  

\bibliographystyle{unsrt}
\bibliography{references}  

\end{document}